\documentclass[[letterpaper, 10 pt, journal, twoside]{IEEEtran}
\usepackage{amsmath,amsfonts,amssymb}
\makeatletter
\newcommand{\removelatexerror}{\let\@latex@error\@gobble}
\makeatother
\usepackage{array}
\usepackage{textcomp}
\usepackage{stfloats}
\usepackage{url}
\usepackage{verbatim}
\usepackage{graphicx}
\usepackage{cite}

\usepackage{dsfont}

\usepackage[caption=false,listofformat=subsimple,labelformat=simple]{subfig}

\usepackage{multirow}
\usepackage{multicol}

\usepackage{xcolor}

\usepackage{hyperref}

\usepackage[normalem]{ulem} 

\usepackage{titlesec} 
\usepackage{indentfirst}
\titlespacing*{\section}{0pt}{0.1em}{0.05em}
\titlespacing*{\subsection}{0pt}{0.1em}{0.05em}
\titlespacing*{\subsubsection}{0pt}{0.1em}{0.05em}
\newcounter{inlinesubsubsection}
\newcommand{\inlinesubsubsection}[1]{%
    \refstepcounter{inlinesubsubsection}%
    \indent\textit{Q\theinlinesubsubsection. #1}:%
}
\begin{document}

\title{GOTPR: General Outdoor Text-based\\Place Recognition Using Scene Graph Retrieval\\with OpenStreetMap}

\author{
Donghwi Jung, Keonwoo Kim, and Seong-Woo Kim
\thanks{Received 14 January 2025; accepted 24 April 2025. Date of publication 8 May 2025; date of current version 19 May 2025. This article was recommended for publication by Associate Editor D. Belter and Editor S. Behnke upon evaluation of the reviewers’ comments.  This work was supported in part by Korea Institute for Advancement of Technology(KIAT) funded by the Korea Government (MOTIE) through HRD Program for Industrial Innovation under Grant P0020536, and in part by the Korean Ministry of Land, Infrastructure and Transport (MOLIT) as the Innovative Talent Education Program for Smart City. The Institute of Engineering Research at Seoul National University provided research facilities for this work. \emph{(Corresponding author: Seong-Woo Kim.)}}
\thanks{The authors are with Seoul National University, Seoul 08826, South Korea {\tt\footnotesize \{donghwijung,knwoo,snwoo\}@snu.ac.kr}.}
\thanks{This article has supplementary downloadable material available at https://doi.org/10.1109/LRA.2025.3568306, provided by the authors.}
\thanks{Digital Object Identifier 10.1109/LRA.2025.3568306}
}

\markboth{IEEE ROBOTICS AND AUTOMATION LETTERS, VOL. 10, NO. 6, JUNE 2025}
{Jung \MakeLowercase{\textit{et al.}}: GOTPR: General Outdoor Text-based Place Recognition Using Scene Graph Retrieval with OpenStreetMap}

\maketitle

\begin{abstract}
We propose GOTPR, a robust place recognition method designed for outdoor environments where GPS signals are unavailable. Unlike existing approaches that use point cloud maps, which are large and difficult to store, GOTPR leverages scene graphs generated from text descriptions and maps for place recognition. This method improves scalability by replacing point clouds with compact data structures, allowing robots to efficiently store and utilize extensive map data. In addition, GOTPR eliminates the need for custom map creation by using publicly available OpenStreetMap data, which provides global spatial information. We evaluated its performance using the KITTI360Pose dataset with corresponding OpenStreetMap data, comparing it to existing point cloud-based place recognition methods. The results show that GOTPR achieves comparable accuracy while significantly reducing storage requirements. In city-scale tests, it completed processing within a few seconds, making it highly practical for real-world robotics applications.
More information can be found at \url{https://donghwijung.github.io/GOTPR_page/}.
\end{abstract}
\begin{IEEEkeywords}
Text-based place recognition, scene graph, OpenStreetMap
\end{IEEEkeywords}
\section{Introduction}
\IEEEPARstart{I}{n} the current era of widespread adoption of autonomous platforms, such as delivery robots~\cite{alverhed2024autonomous}, it is necessary to communicate the caller's current scene to the robot to summon it. A user provides the environmental information to the robot, which then determines the scene using its stored map data.
Various approaches are available for transmitting spatial information to the robot, including capturing the surroundings with a smartphone’s camera or LiDAR and sending the scene as images or point clouds \cite{miao2024scenegraphloc,woo2024context,xia2021soe}. However,
among the various ways information can be conveyed to robots, verbal communication is one of the most natural approaches for humans~\cite{nikolaidis2018planning}. In this context, some researches have explored text-based place recognition, where natural language is used to communicate information to robots\cite{kim2024e2map,hong2019textplace,kolmet2022text2pos,wang2023text,xia2024text2loc,shang2024mambaplace,liu2025text,chen2024scene}.\\
\indent Unlike methods that rely on physically present scene texts~\cite{hong2019textplace}, other approaches~\cite{kolmet2022text2pos,wang2023text,xia2024text2loc,shang2024mambaplace,liu2025text} utilize user-provided text inputs. They embed both the text and pre-segmented point cloud maps using contrastive learning~\cite{le2020contrastive}, and perform place recognition by finding the most similar embedding pair.
While these methods achieve high retrieval accuracy, they rely on pre-segmented point cloud maps \cite{liao2022kitti}, which have several limitations. The creation of semantic point cloud maps is time-consuming, and their large data size makes it impractical for robots to store map data for extensive areas. These limitations hinder the application of \cite{kolmet2022text2pos,wang2023text,xia2024text2loc,shang2024mambaplace,liu2025text} in real-world robots.\\
\indent To address these issues, Where am I?\cite{chen2024scene} leveraged scene graphs\cite{armeni20193d} to model map data, effectively reducing storage demands. The method was evaluated in indoor settings using the 3DSSG~\cite{wald2020learning} and the ScanScribe datasets~\cite{zhu20233d}. A transformer architecture was employed to compute joint embeddings for text graphs and scene graphs, and their similarities were analyzed to identify the most analogous graph pair for place recognition. Representing map data through scene graphs allowed for a more compact storage requirement in their approach. However, the method was tested exclusively in indoor environments, leaving outdoor environments unaddressed. Outdoor environments present additional challenges due to the larger scope of scenes to be compared. Calculating joint embeddings for all scene graphs, as in \cite{chen2024scene}, is computationally expensive, making real-world application in robots infeasible. To overcome this computational burden, in this paper we propose a method that first extracts matching candidates from the vectorDB \cite{wang2021milvus} using text embeddings, followed by computing joint embeddings with these candidates, thereby reducing overall processing time.\\
\indent To further increase the practical applicability of our method, we utilize OpenStreetMap (OSM) \cite{haklay2008openstreetmap} for outdoor map data. OSM is an open dataset containing information on outdoor spaces globally, making it versatile for use in various environments. Additionally, OSM’s categorization of data into nodes, ways, and relations makes it well-suited for conversion into scene graphs \cite{tully2016generating}. Unlike previous study~\cite{zipfl2022towards}, which represent scene graphs as numerical spatial relationships based on distances and angles, our approach integrates text-based information to solve the text-based place recognition problem, finding the matching scene based on text queries.\\
\indent The contributions of this paper are as follows:
\vspace{-0.1cm}
\begin{itemize}
\item We propose GOTPR, a method leveraging scene graphs to reduce map data storage requirements and maintain consistent algorithm speed even as the number of frames increases, making it suitable for real-world applications.
\item The proposed GOTPR utilizes OSM data to enable place recognition in various outdoor environments without the need for preprocessing steps such as map generation, thereby enhancing its practicality.
\item We validate the proposed approach in city-scale environments, demonstrating high accuracy, real-time performance, and minimal map data storage requirements, confirming its applicability to real-world robotic systems.
\end{itemize}
\begin{figure*}[t]
    \centering
    \includegraphics[width=0.99\textwidth]{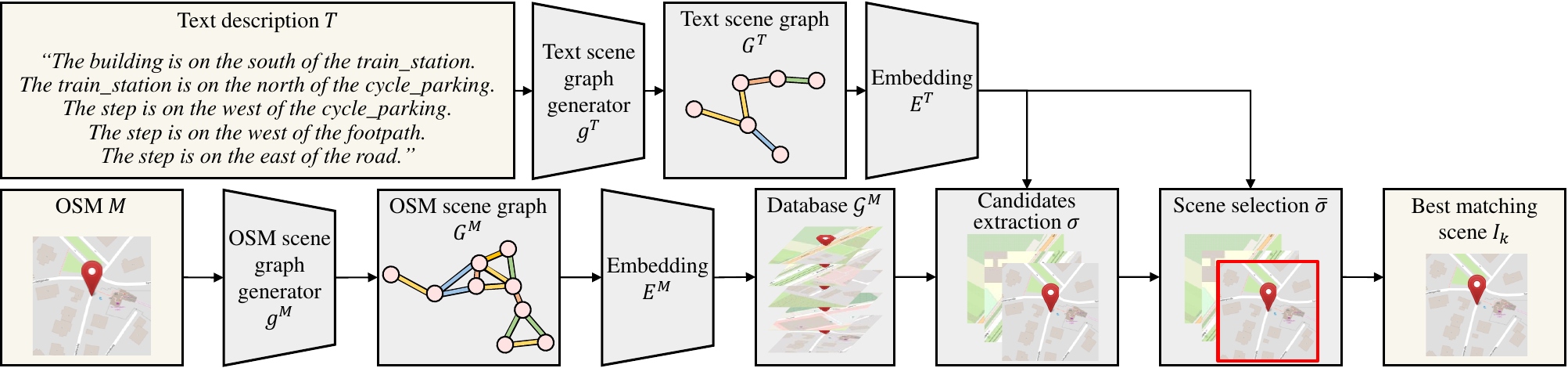}
    \caption{Process of GOTPR. The method consists of three sequential steps: 1) Scene graph generation, 2) Scene graph candidates extraction, 3) Scene graph retrieval and scene selection. The input data are a query text description and OSM. The output is the matching scene id.}
    \label{fig:process}
\end{figure*}
\section{Related Works}
\subsection{Point cloud map-based}
Existing methods leveraging text for place recognition include \cite{kolmet2022text2pos,wang2023text,xia2024text2loc,shang2024mambaplace,liu2025text}. These studies employ the KITTI360Pose dataset\cite{kolmet2022text2pos}, and focus on place recognition by comparing the similarity between point cloud maps and textual descriptions. Each study differs slightly depending on the methodologies used for comparing text and point cloud maps. First, Text2Pos\cite{kolmet2022text2pos} utilized an bidirectional Long Short-Term Memory (LSTM)~\cite{hochreiter1997long}, to identify similar scenes.
Building on this work, RET~\cite{wang2023text} introduced the relation enhanced transformer network, which improved performance by employing self-attention mechanism to compare text and point cloud maps.
Subsequently, Text2Loc\cite{xia2024text2loc} improved learning performance by utilizing a hierarchical transformer with max-pooling (HTM) based on contrastive learning, unlike RET\cite{wang2023text}.
Furthermore, MambaPlace\cite{shang2024mambaplace} enhanced the model’s reasoning ability over a broader range by leveraging a Mamba-based State Space Model (SSM)\cite{gu2023mamba} with attention mechanisms. Their approach currently represents the state-of-the-art performance for text-based place recognition using the KITTI360Pose dataset.
MNCL~\cite{liu2025text} trains cross-modal encoders using multi-level negative contrastive learning to identify similar pairs.
However, these methods rely on point cloud maps for place recognition, which poses challenges such as storage requirements and scalability to larger spaces, limiting their practical applicability in real-world robotic systems.
\subsection{Scene graph-based}
In contrast, Where am I?~\cite{chen2024scene} employed 3D scene graphs~\cite{armeni20193d}, for the text-based place recognition method. The research assessed the model's performance using indoor scene graph datasets, including 3DSSG~\cite{wald2020learning} and ScanScribe~\cite{zhu20233d}. In addition, word2vec~\cite{church2017word2vec} was applied to process node and edge labels, generating both text and scene graphs. These graphs were then embedded using a graph transformer network with message passing, as proposed by~\cite{ijcai2021p214}, and similarity scores were calculated through a Multi-Layer Perceptron (MLP) to identify the most similar scenes. By representing scene data as scene graphs, their method addresses the challenges of existing text-based place recognition techniques~\cite{kolmet2022text2pos,wang2023text,xia2024text2loc,shang2024mambaplace,liu2025text}, which rely on large-scale point cloud map data. However, the scope of Where am I?~\cite{chen2024scene} is confined to indoor environments.\\
\indent In this paper, we extend the scope of text-based place recognition utilizing scene graph retrieval to outdoor environments, enabling it to operate in more diverse settings. To achieve this, we incorporate OSM, the graph Transformer network, and scene graph embedding data to perform an initial extraction of comparison candidates. As a result, the proposed method features compact scene data storage, fast processing speed, and high accuracy, establishing its significance as a real-time, high-performance city-scale outdoor place recognition solution.
\begin{figure*}[t]
    \centering
        \subfloat[OSM scene graph.]{
        \includegraphics[width=0.99\textwidth]{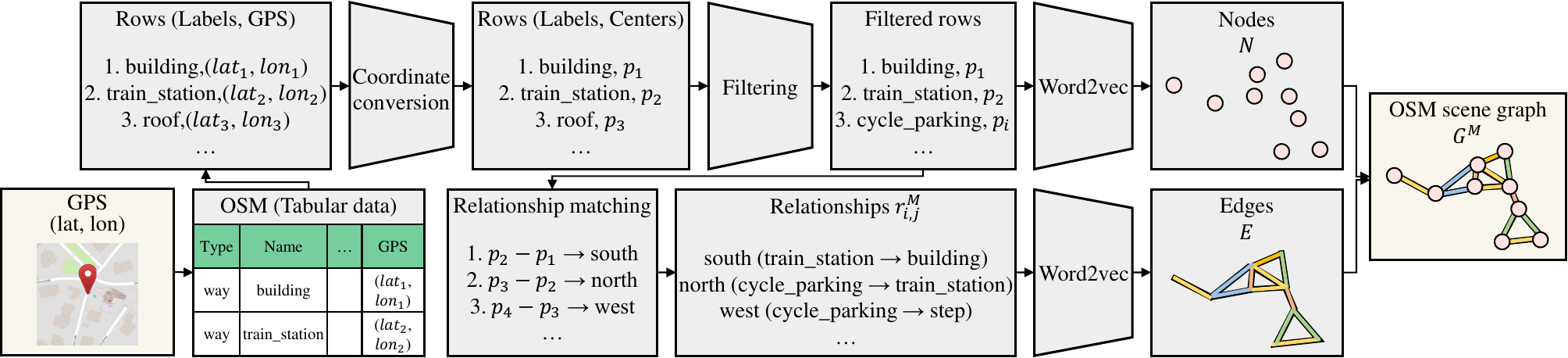}
        }\\
        \vspace{-0.3cm}
        \subfloat[Text scene graph.]{
    	  \includegraphics[width=0.99\textwidth]{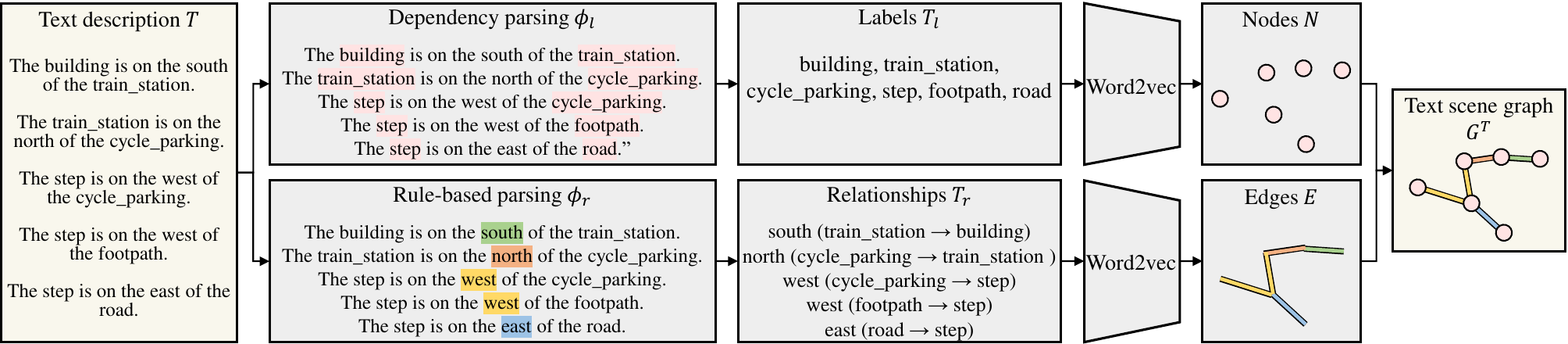}
        }
    \caption{Process of scene graph generation. (a) OSM scene graph generation. (b) Text scene graph generation.}
    \label{fig:scene_graph_generation}
\end{figure*}
\section{Methods}
The process of our proposed method is illustrated in Fig.~\ref{fig:process}.
\subsection{Scene graph generation}\label{method:scene_graph_generation}
First, the input data, consisting of text and OSM, is converted into scene graphs. The scene graph generation process is depicted in Fig.~\ref{fig:scene_graph_generation}.
\begin{gather}
    G=\{N,\;E\},\quad N=\{n\},\quad E=\{e\},
\end{gather}
where $G$ refers to a scene graph. $n$ and $e$ denote the nodes and edges that constitute the scene graph $G$, respectively. At this point, the nodes and edges correspond to the labels and relationships shown in Fig.~\ref{fig:scene_graph_generation}. $N$ and $E$ represent the sets of $n$ and $e$, respectively.\\
\indent The extraction of node labels from the text description is performed using dependency parsing~\cite{nivre2010dependency}. To distinguish duplicate node labels within the same scene, identifiers appended to the labels are detected and removed during the node label extraction process using a rule-based method. For the extraction of edge relationships, we employ a rule-based approach, where a predefined set of words is established, and words present in this set are identified in the sentence and used as edges. Using the extracted node labels and edge relationships, a text scene graph is constructed.
\begin{gather}
    T_l = \phi_l(T),\quad T_r = \phi_r(T),\\\label{eqn:scene_graph_generation}
    G^T=g^T(T_l, T_r),
\end{gather}
where $T$ represents the text description that depicts the scene. $\phi_l$ and $\phi_r$ are functions that extract labels and spatial relationships from the text description, respectively. $T_l$ and $T_r$ represent the sets of extracted labels and spatial relations, respectively. $G^T$ refers to the scene graph generated from the text. Additionally, $g^T$ denotes the function that generates $G^T$ using two textual components.\\
\indent In the processing of OSM data, node labels are assigned by extracting element names from the loaded dataset within a specified distance from a reference GPS point. The corresponding GPS coordinates are then transformed into the East-North-Up (ENU) coordinate system. Each node’s position is determined by identifying the nearest point, defined in ENU coordinates, within an element’s polygon relative to the reference location. Using these node positions, spatial relationships are analyzed, and edges are established between nodes when their Euclidean distance falls below a predefined threshold.
The positional difference between nodes is represented as a vector, which is used to determine the relationship indicated by the edge. Additionally, the edge relationships in both the text and OSM scene graphs are expressed using four directional components: North, South, East, and West. These four components correspond to the rule-based word set used during the scene graph generation process in Eq.~\eqref{eqn:scene_graph_generation}. By representing relationships in different scene graphs using a common framework, the edge relationships in scene graphs derived from two distinct data sources—text data and OSM data—are expressed in the same coordinate system, enabling direct graph-to-graph comparisons. The scene graph generation process can be mathematically formulated as follows:
\begin{gather}
    r^M_{i,j}=\begin{cases}
    North,&\text{if}\;\|\Delta p\|_x < \|\Delta p\|_y, (\Delta p)_y > 0 \\
    South,&\text{else if}\;\|\Delta p\|_x < \|\Delta p\|_y, (\Delta p)_y < 0 \\
    East,&\text{else if}\;\|\Delta p\|_x > \|\Delta p\|_y, (\Delta p)_x > 0 \\
    West,&\text{else if}\;\|\Delta p\|_x > \|\Delta p\|_y, (\Delta p)_x < 0 \end{cases},\label{eqn:nsew}
\end{gather}
where $p$ represents the position of an object in the global coordinate system, and $\Delta p$ is the vector calculated as the positional difference between two objects, $p_j$ and $p_i$. This vector is used to compute the spatial relation $r^M_{i,j}$ as in Eq.~\eqref{eqn:nsew}. $\|\cdot\|_{x,y}$ denotes the size of elements associated with $x$ or $y$ within the vector.
\begin{gather}
    G^M=g^M(M),\quad ^\forall M\in\mathcal{M},\\
    \mathcal{G}^M=\{G^M\},
\end{gather}
where $M$ is the submap cropped from the OSM based on a predefined distance. $\mathcal{M}$ represents the set of $M$, corresponding to the entirety of the OSM. $G^M$ denotes to the scene graph generated from the OSM. $g^M$ indicates the function that generates $G^M$ using the map data. $\mathcal{G}^M$ refers the set of $G^M$, constituting the scene graph database for the entire OSM.
\begin{figure*}[t]
    \centering
    \includegraphics[width=0.99\textwidth]{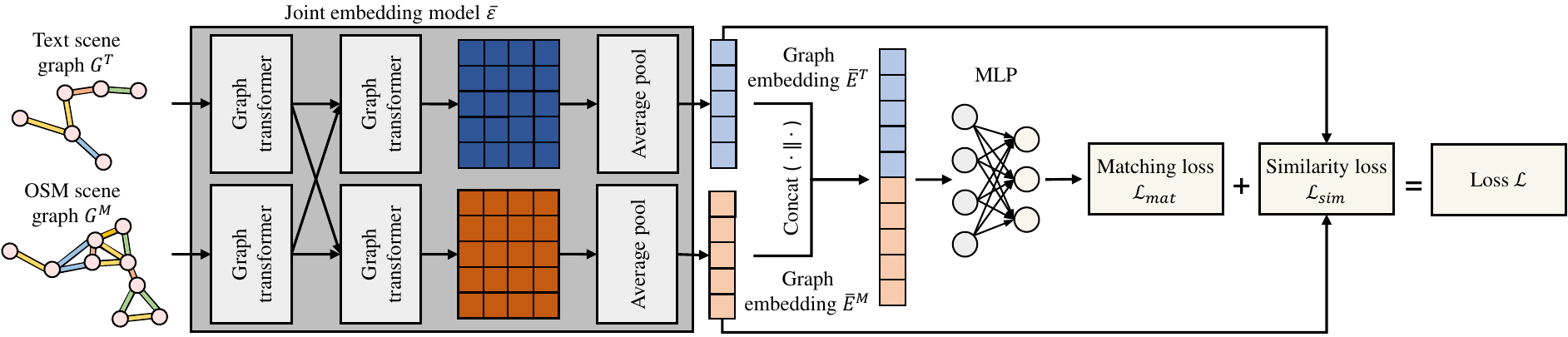}
    \caption{Process of scene graph retrieval. Input of the process are query text scene graph and the extracted OSM scene graph candidates. And the output is the selected top-k scene id. The joint embedding model consists of multiple GPS convolution layers with self and cross modules.}
    \label{fig:scene_graph_retrieval}
\end{figure*}
\subsection{Similar scene graph candidates extraction}\label{method:scene_graph_candidates}
To efficiently compare similarities between scene graphs stored in the database, converted from OSM into scene graphs, we first extract a set of candidate graphs that are expected to have high similarity. For this purpose, we utilize the graph embedding module for text and OSM scene graphs introduced in Sec.~\ref{method:scene_graph_retrieval}. Using this module, we compute the embedding data for OSM scene graphs and store them in a vectorDB.
\begin{gather}
    E^T=\varepsilon(G^T),\\
    E^M_i=\varepsilon(G^M_i),\quad ^\forall i\in\mathcal{I},
\end{gather}
where $\varepsilon$ denotes a function that generates a graph embedding from either a text graph or a scene graph as input. $E$ refers to the graph embedding produced by $\varepsilon$. $\mathcal{I}$ represents the set of all scene ids stored in the database.\\
\indent Then, we identify the embedding data most similar to the query text scene graph's embedding data. The similarity between embedding vectors is calculated using cosine similarity.
\begin{equation}
    S_i=\sigma(E^T,E^M_i)=\frac{E^T\cdot E^M_i}{\|E^T\|\cdot\|E^M_i\|},
\end{equation}
where $\sigma$ refers to a function that calculates the similarity between two scene graph embeddings. $S$ corresponds to the similarity value produced as a result of $\sigma$. $\|\cdot\|$ is the size of the vector. Subsequently, we extract the OSM scene ids with high similarities.\\
\indent Using the scene ids with high similarities, the corresponding candidate map scene graphs are extracted from the map scene graphs stored in the vectorDB.
\begin{equation}
\mathcal{C}=\{G^M_i\}=\psi(G^T,\mathcal{G}^M),\quad ^\forall i\in \mathbf{I}_n,\label{eqn:candidates_extraction}
\end{equation}
where $\mathcal{C}$ is the set composed of map scene graphs $G^M_i$ that exhibit high similarity with the text scene graph $G^T$. $\psi$ represents the function used to extract $\mathcal{C}$. $\mathbf{I}_n$ represents the set of scene ids selected as the output of the \emph{argtop-$n$} function.
\subsection{Scene graph retrieval and scene selection}\label{method:scene_graph_retrieval}
As illustrated in Fig.~\ref{fig:scene_graph_retrieval}, we employ a joint embedding framework, leveraging a graph transformer network, to generate embeddings for scene graphs, referring to \cite{chen2024scene}. Based on the model, the similarity between these graph embeddings is calculated, and the scene with the highest similarity is determined as the one corresponding to the current query text description. For the graph transformer network, we configure the network in the form of a joint embedding model comprising these graph transformer layers~\cite{ijcai2021p214}. The reason for utilizing this layer is to enable each node to effectively integrate information from its neighboring nodes by leveraging a Multi-Head Attention~\cite{vaswani2017attention}-based message passing mechanism.
\begin{gather}
    \bar{E}^T,\bar{E}^M_i=\bar{\varepsilon}(G^T,G^M_i),\quad ^\forall G^M_i\in \mathcal{C},\\
    \bar{S}_i=\bar{\sigma}(\bar{E}^T,\bar{E}^M_i),
\end{gather}
where $\bar{\varepsilon}$ denotes the joint embedding model that transforms a text graph and a scene graph. $\bar{E}$ represents the joint embedding of the text graph and scene graph generated as the output of $\bar{\varepsilon}$. $\bar{S}$ is the cosine similarity between two joint embedding data. $\bar{\sigma}$ indicates the function that calculates $\bar{S}$.\\
\indent Finally, the scene id is predicted by applying the \emph{argtop-$n$} operation to the cosine similarity scores.\\
\indent The constructed scene retrieval network is trained using a loss function composed of two components. First, the matching probability loss is computed based on the matching probability predicted by an MLP for two scene graph embeddings. Second, the cosine similarity loss is defined as the sum of the cosine similarity values directly calculated between the two embedding vectors. The joint embedding network is trained using this combined loss.
\begin{gather}
    \mathcal{L}_{mat}=MLP\left(\bar{E}^T\|\bar{E}^M_i\right),\quad \mathcal{L}_{sim}=\bar{S}_i,\\
    \mathcal{L}=\mathcal{L}_{mat}+\mathcal{L}_{sim},
\end{gather}
where $MLP$ refers to the MLP component within the scene graph retrieval network architecture. $\mathcal{L}$ denotes the loss function of the scene graph retrieval network. $\mathcal{L}_{mat}$ and $\mathcal{L}_{sim}$ are the two components of $\mathcal{L}$, representing the matching probability loss and cosine similarity loss, respectively. $(\,\cdot\,\|\,\cdot\,)$ indicates the concatenation of two vectors.
\begin{figure*}[t]
    \centering
    \includegraphics[width=0.99\textwidth]{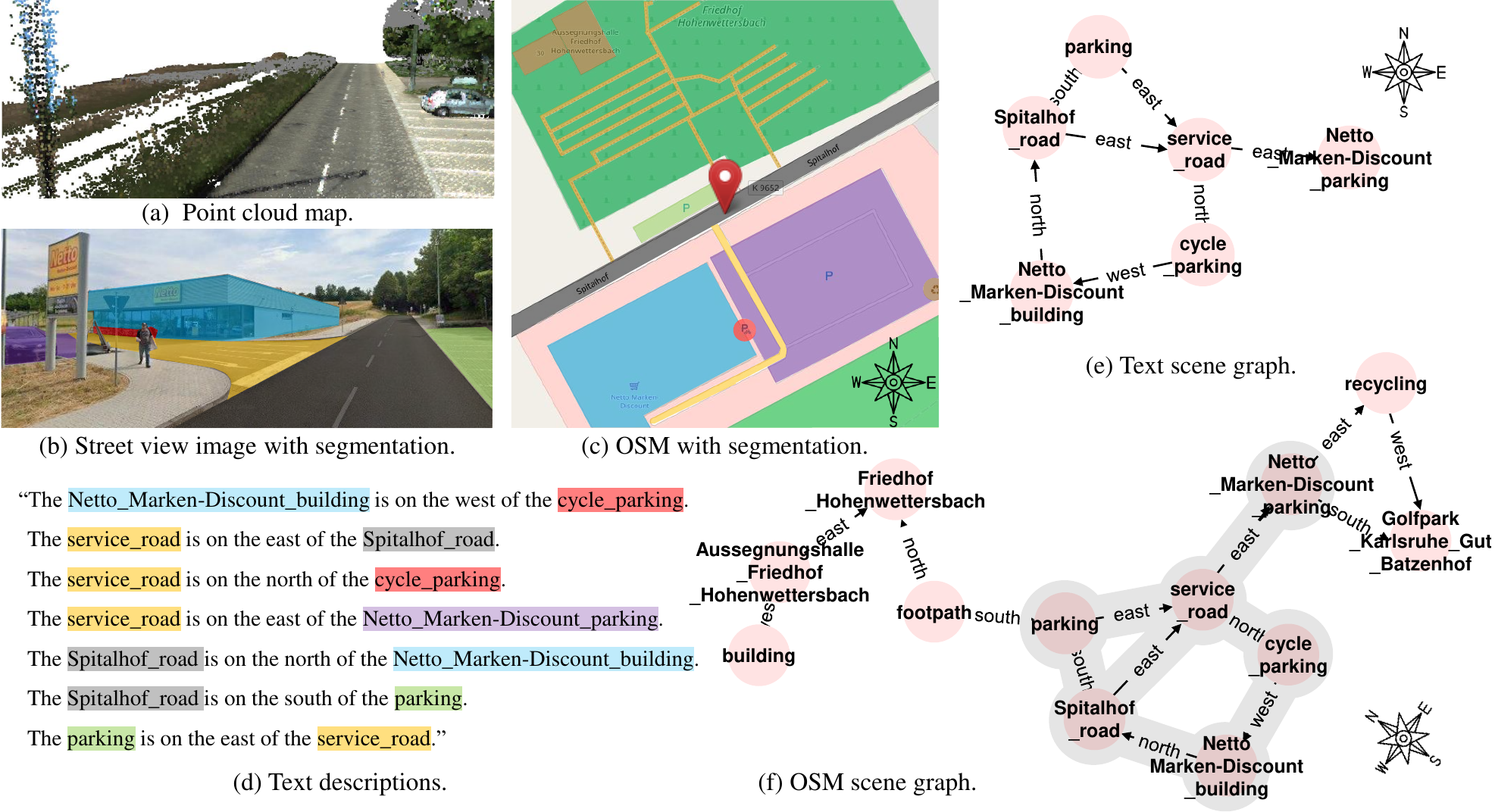}
    \vspace{-0.2cm}
    \caption{Examples of the experimental data are provided. The data were generated based on GPS coordinates 48.964117, 8.472481.
    (a) and (b) are presented to facilitate comparison and understanding rather than actual data used in GOTPR.
    In addition, The segmentations in (b) and (c) are included solely for visualization purposes to aid understanding. Moreover, the gray area in (f) represents the overlapping region with (e).}
    \label{fig:experiments_results}
\end{figure*}
\section{Experiments}
\subsection{Baselines and metrics}\label{exp:details}
In this paper, to evaluate the performance of the proposed GOTPR, we adopted existing text-based place recognition methods, including Text2Pos\cite{kolmet2022text2pos}, RET\cite{wang2023text}, Text2Loc\cite{xia2024text2loc}, MambaPlace\cite{shang2024mambaplace}, and Where am I?\cite{chen2024scene}, as baselines.
MNCL~\cite{liu2025text} was excluded from the baseline as its code has not been released.
Among these, Text2Pos, RET, Text2Loc, and MambaPlace were validated using the KITTI360Pose dataset. For a comparative analysis with these baselines, as shown in Table~\ref{table:accuracy}, GOTPR converted the segmented point cloud maps from the KITTI360Pose dataset into scene graphs. Additionally, the GPS coordinates corresponding to the poses in the KITTI360Pose dataset were utilized to generate submaps from OSM, which were then converted into scene graphs to serve as map data.
To evaluate the performance of the proposed algorithm, we used Retrieval Recall as the evaluation metric. This metric calculates the ratio of correctly matched true positive OSM scenes stored in the database among all tested text queries, expressed as a value between 0 and 1. Notably, each text query has exactly one true positive OSM scene. The matching success is determined based on the true positive OSM scene id for each text query. In this process, translation and rotation are not considered. As formalized in Eq.~\eqref{eqn:candidates_extraction}, the final predicted scene is identified by selecting the top $k$ candidates with the highest similarity, and Retrieval Recall is computed by checking whether the true positive scene is included among these candidates.
In this paper, $k$ was set to three different values—1, 3, and 5—by referencing prior text-based place recognition studies~\cite{kolmet2022text2pos,wang2023text,xia2024text2loc,shang2024mambaplace}.
\subsection{Implementation}\label{exp:implementation}
Following prior studies\cite{srivamsi2023cosine}, the Milvus\cite{wang2021milvus} was employed as the vectorDB for storing the map scene graphs. The hardware configuration for this paper consisted of an Intel Xeon Gold 6140 2.30 GHz CPU and a 12 GB NVIDIA TITAN Xp GPU. The hardware specifications are comparable to those of commercially available edge computing devices, suggesting that the experimental results can be applied to real-world robotic scenarios.
In the KITTI360Pose dataset, each map includes a minimum of six objects, with each object associated with a corresponding text description that reflects its pose. Consequently, the resulting map scene graph contains at least six nodes, and the text scene graph comprises six nodes. Therefore, in this paper, only OSM data that yielded six or more elements when retrieved using GPS coordinates was utilized, ensuring that each OSM scene graph also contains at least six nodes, consistent with the structure of the KITTI360Pose map scene graphs. The text scene graphs were constructed to include exactly six nodes. However, unlike KITTI360Pose, the text descriptions were generated based on spatial relationships among objects, which occasionally resulted in more than six descriptions. No constraint was imposed on the number of object categories or types in either the OSM or text scene graphs. Each OSM scene graph was created by querying OSM data within a 50-meter radius from a given GPS location.
\begin{table*}[!t]
\caption{comparison between our method and existing text-based place recognition methods.}
\centering
\def\arraystretch{1.5}%
\setlength{\tabcolsep}{12pt}
\label{table:accuracy}
\begin{tabular}{c|c|ccccccc}
    \hline
    \multicolumn{1}{c}{}&\multicolumn{1}{c}{}&\multicolumn{6}{c}{Submap Retrieval Recall $\uparrow$}\\\cline{3-8}
    \multicolumn{1}{c}{}&\multicolumn{1}{c}{}&\multicolumn{3}{c}{Validation Set}&\multicolumn{3}{c}{Test Set}\\\cline{3-8}
    \multicolumn{1}{c}{Map data}&\multicolumn{1}{c}{Methods} & $k=1$& $k=3$ & $k=5$& $k=1$ & $k=3$ & $k=5$\\\hline
    \multirow{8}{*}{KITTI360Pose}&Text2Pos~\cite{kolmet2022text2pos} &0.14&0.28&0.37&0.12&0.25&0.33\\
    &RET~\cite{wang2023text} &0.18&0.34&0.44&-&-&-\\
    &Text2Loc~\cite{xia2024text2loc} &0.32&0.56&0.67&0.28&0.49&0.58\\
    &MambaPlace~\cite{shang2024mambaplace} &0.35&0.61&0.72&0.31&0.53&0.62\\\cline{2-8}
    &GOTPR (ours; rule-based; observer-centric) &0.39&0.68&0.80&0.26&0.51&0.62\\
    &GOTPR (ours; Transformer; observer-centric)&0.27&0.48&0.58&0.20&0.37&0.45\\\cline{2-8}
    &GOTPR (ours; rule-based) &0.39&0.68&0.80&0.24&0.52&0.66\\
    &GOTPR (ours; Transformer) &0.36&0.58&0.67&0.30&0.49&0.57\\\hline
    \multirow{2}{*}{OSM}&GOTPR (ours; rule-based) &0.24&0.48&0.57&0.29&0.50&0.58\\
    &GOTPR (ours; Transformer) &0.23&0.43&0.53&0.22&0.39&0.48\\\hline
\end{tabular}
\end{table*}
\subsection{Research questions}\label{exp:research_questions}
\indent\emph{Q1.} Can place recognition performance remain comparable even when the environment is represented with simplified data structures such as scene graphs, instead of precise data like point clouds?
\emph{Q2.} Does the scene graph generated from OSM achieve comparable overall algorithm performance to the scene graph generated from segmented point cloud maps?
\emph{Q3.} Does the proposed transformer network-based approach outperform the rule-based scene graph retrieval method, in terms of performance or provide advantages in speed?
\emph{Q4.} Compared to a brute-force retrieval method that evaluates all scenes, how much does the pre-selection of similar scene candidates improve speed, and does it maintain comparable performance?
\emph{Q5.} When scene data is converted from point clouds to scene graphs, how much storage efficiency is gained? Is it feasible for application in real-world robots?
\emph{Q6.} When applied to city-scale data, do accuracy, speed, and data storage requirements produce results that are practical for real-world robotic applications?
\emph{Q7.} Does the quality of the text description affect the performance of GOTPR?
\subsection{Evaluation}
The data examples used in the experiments are illustrated in Fig.~\ref{fig:experiments_results}. These examples include text descriptions and GPS coordinates serving as input data for GOTPR, the corresponding OSM data retrieved based on these inputs, and the resulting text and map scene graphs generated from this data. To further enhance comprehension, the point cloud map obtained from the KITTI360pose dataset~\cite{kolmet2022text2pos} and the street view image\cite{google_street_view} of the area are also included. It should be noted that the segmentations shown in Fig.~\ref{fig:experiments_results}(b) and \ref{fig:experiments_results}(c) are utilized solely for visualization purposes to facilitate understanding. Through the use of these data, the experiments aim to address the research questions outlined earlier in Sec.~\ref{exp:research_questions}. The results of these validations are as follows:\\
\inlinesubsubsection{Scene graph-based accuracy}
The performance of methods leveraging scene graphs is shown in Table~\ref{table:accuracy}. As described in Sec.~\ref{exp:details}, place recognition performance was evaluated by converting map data from the KITTI360Pose dataset into scene graphs. In the KITTI360Pose dataset, the direction was computed based on the relative positional differences between the observer and the object. On the other hand, GOTPR does not include the observer as a node in the scene graph, so using the direction information between the observer and the object as an edge in the scene graph is not appropriate. Therefore, we calculated the direction by applying the positional data of objects to Eq.~\eqref{eqn:nsew} and utilized it as edges in the scene graph for this experiment. Additionally, to perform a fair performance comparison with existing studies, we directly used the KITTI360Pose dataset, which includes the observer in the scene graph, and created the scene graph based on the positional relationship between the observer and objects, then showed the place recognition results. The results indicate that whether the scene graph was created around the observer or not, the performance difference between using scene graph-based map data and semantically segmented point cloud maps was not significant. This suggests that the scene graph-based map data used in GOTPR contains sufficient information to be effectively utilized for place recognition.\\
\inlinesubsubsection{OSM-based accuracy}
The place recognition performance of GOTPR using scene graphs generated from OSM is presented in Table~\ref{table:accuracy}. The place recognition performance achieved using the OSM scene graph is slightly lower or comparable to the performance obtained when directly utilizing the point cloud map or its transformation into a scene graph. This demonstrates that it is unnecessary to rely on pre-generated, complex data structures such as point clouds for map data.
An additional advantage of OSM data lies in its continuous updates, which allow it to better reflect real-world changes over time. In contrast, point cloud maps become outdated when the environment evolves after data acquisition, potentially leading to place recognition failures, as demonstrated in Fig.~\ref{fig:experiments_results}. For example, a location initially mapped as an empty lot may become unrecognizable if a building is later constructed. Nevertheless, challenges arise from the open and crowdsourced nature of OSM, which can result in incomplete or inconsistent labeling. To address these issues, we explored alternative labeling strategies, including leveraging supplementary element data to generate more descriptive node labels when original tags were ambiguous or insufficient, such as replacing vague labels like \textit{Yes} with more specific ones such as \textit{House} or \textit{Office}.\\
\inlinesubsubsection{Network and rule-based}
The comparison of different scene graph retrieval methods and network architectures is presented in Table~\ref{table:accuracy}. For the rule-based approach, Graph Edit Distance (GED) was utilized, which is a widely used method for measuring graph similarity as mathematically proven in \cite{sanfeliu1983distance} and detailed in \cite{gao2010survey}. The results indicate that the rule-based approach using GED achieves the highest accuracy on average. However, we conducted experiments on GOTPR under three conditions: 1) rule-based, 2) network-based on CPU, and 3) network-based on GPU. The results showed that the number of text-to-scene pairs processed per second was 11.05, 80.44, and 152.96, respectively. This demonstrates that our proposed transformer network-based method, when executed on a GPU, is approximately 13.84 times faster than the rule-based approach.
Considering the need for real-time operation, network-based methods are more practical.\\
\inlinesubsubsection{Candidates extraction}
As shown in the Table~\ref{table:ablation_study_processing_time}, 
we conducted experiments to evaluate the accuracy and processing speed of our method when using either the vectorDB-based candidates extraction in Sec.~\ref{method:scene_graph_candidates} or the scene retrieval network shown in Fig.~\ref{fig:scene_graph_retrieval}, or both. In these experiments, the value of \emph{argtop-$n$}, which represents the number of candidates extracted in Eq.~\eqref{eqn:candidates_extraction}, was set to 10. For the experimental data, we used the \emph{2013\_05\_28\_drive\_0010\_sync} sequence from the validation dataset of the scene graphs created from the KITTI360Pose dataset, as shown in Table~\ref{table:accuracy}.\\
\indent The experimental results show that the highest accuracy is achieved when only scene graph retrieval is used. This is because, in candidates extraction, only the top-$n$ candidates, which are predefined, are extracted. If the actual matching scene is not included in these top-$n$ candidates, there is no way to correct this in the subsequent scene retrieval process. For this reason, when both candidates extraction and scene graph retrieval are used together, the accuracy is lower than when only scene graph retrieval is used, but this difference can be reduced by increasing the value of $n$ in the top-$n$.\\
\indent Additionally, we found that, without candidates extraction, it takes approximately 1,850 milliseconds to process a single text scene graph. This is nearly 2 seconds, and when applied in a real-time application, it could cause delays. On the other hand, when candidates extraction is applied, we can find the matching pair within a short time of either 0.43 or 37.11 milliseconds. This difference in processing speed becomes even more pronounced in large-scale environments, such as city-level scenes, where the number of scene graphs to be compared is large. Therefore, by using candidates extraction to select appropriate candidates and then performing scene graph retrieval on those candidates, higher performance and faster processing speed can be achieved.\\
\begin{table}[!t]
\caption{comparison of accuracy and processing time.\\(C.E.:Candidates Extraction, S.G.R.:Scene Graph Retrieval)}
\centering
\def\arraystretch{1.5}%
\setlength{\tabcolsep}{8pt}
\label{table:ablation_study_processing_time}
    \begin{tabular}{ccccc}
    \hline
    &\multicolumn{3}{c}{Accuracy}&Time ($ms$)\\
    \cline{2-5}
    &$k=1$& $k=3$ & $k=5$&\\\hline
    C.E.&0.24&0.42&0.50&\textbf{0.43 $\pm$ 0.68}\\
    S.G.R.&\textbf{0.36}&\textbf{0.58}&\textbf{0.67}&1,849.69 $\pm$ 98.85\\
    C.E. and S.G.R.&0.33&0.51&0.57&37.11 $\pm$ 10.03\\\hline
    \end{tabular}
\end{table}
\begin{table}[!t]
\caption{performance on city-scale data.}
\centering
\def\arraystretch{1.5}%
\setlength{\tabcolsep}{13.5pt}
\label{table:city_scale_place_recognition}
\begin{tabular}{cccc}
\hline
\multicolumn{1}{c}{}& Accuracy&Speed&Size\\
\multicolumn{1}{c}{}&(Top $k$ = 1 / 3 / 5)&($s/iter$)&($MB$)\\\hline
Karlsruhe&0.28 / 0.44 / 0.52&0.15&15.44\\ 
Sydney&0.23 / 0.38 / 0.49&0.13&34.31\\
Toronto&0.22 / 0.36 / 0.44&0.29&38.74\\\hline 
\end{tabular}
\end{table}
\inlinesubsubsection{Scene graph-based storage size}
We also examined the changes in storage requirements when using scene graphs for map data. The size of the map data was measured based on the \emph{2013\_05\_28\_drive\_0003\_sync} scene from the KITTI360~\cite{liao2022kitti} and KITTI360Pose~\cite{kolmet2022text2pos} datasets. In this case, the point cloud maps and scene graphs were constructed using static objects from the overall set of objects. This choice is motivated by the problem we aim to address, which is robust place recognition that operates independently of time variations. According to the KITTI360Pose dataset, the scene is divided into 447 cells, each forming a square with a side length of 30 $m$. Summing the areas of these cells gives a total area of approximately 0.40 $km^2$. Comparing the storage requirements of the map data by data type, the storage requirements for point clouds and scene graphs are 5,546.19 $MB$ and 0.71 $MB$, respectively. This indicates that scene graphs require approximately 7,846 times less storage space than point clouds.\\
\indent If we assume the map area is expanded to encompass the entire city of Karlsruhe, where the KITTI360 dataset was collected, the point cloud map would require approximately 2.29 $TB$ of storage capacity for the entire 173.5 $km^2$ area. However, a storage requirement of more than 2 $TB$ for map data would be a burdensome condition for general autonomous navigation robots equipped with low-level hardware specifications. In contrast, when the map is represented using a scene graph, an approximate calculation based on the proportional relationship with the point cloud map’s size shows that the storage requirement would be about 0.30 $GB$, which is less than 1 $GB$. This demonstrates that, unlike point clouds, scene graph-based map data can be effectively utilized on general robots.\\
\inlinesubsubsection{City-scale}
Finally, assuming GOTPR’s application to real-world robots, we evaluated its accuracy, speed, and storage requirements using city-scale map data. The results, shown in Table~\ref{table:city_scale_place_recognition}, include tests in three regions: Karlsruhe, Sydney, and Toronto. These regions exhibit a range of characteristics, encompassing both larger and smaller cities, thereby facilitating a thorough evaluation of the performance of the proposed algorithm.
For the experiment, we assumed that each city was segmented into a rectangular shape on the OSM map as close as possible to the city boundaries. We then calculated the latitude and longitude ranges corresponding to the corner points of the rectangle.\\
\indent The latitude and longitude ranges for Karlsruhe, Sydney, and Toronto are (48.94, 8.28) to (49.09, 8.54), (-33.92, 150.92) to (-33.77, 151.17), and (43.67, -79.55) to (43.77, -79.30), respectively. Using these GPS ranges, we divided each city into a fixed number of segments, evenly distributing the GPS coordinates to generate a uniform distribution, from which we loaded the corresponding OSM data. As mentioned earlier in Sec.~\ref{exp:implementation}, areas with insufficient OSM data were filtered out, and only the remaining GPS coordinates with matching OSM data were used. The filtering criterion for OSM data was based on the number of OSM elements, with a threshold of six, and the distance threshold used in the OSM query was set to 50~$m$.\\
\indent The areas span 173.5, 12,145, and 630.2 $km^2$, respectively, and were divided into 1,000, 486, and 479 cells, with each cell generating text and map scene graphs for place recognition. The results show a high prediction accuracy exceeding 0.44, with an average processing time of less than 0.3 seconds, suitable for real-time robotic applications. Additionally, the map data size remains under 40 $MB$, making it feasible to store on small storage devices typically used in robots.\\
\inlinesubsubsection{Text description quality}
To assess the influence of text description quality on the performance of GOTPR, we compared two approaches for generating scene descriptions. The first approach followed the original GOTPR pipeline and used OSM data to extract structured text descriptions based on object labels and spatial relationships. The second approach involved human annotators who described street view images retrieved through the Google Street View Static API~\cite{google_street_view} using natural language. Annotators were provided with heading angle information to facilitate the use of cardinal directions such as north, south, east, and west. These free-form descriptions were then converted into the structured text format used by GOTPR using the ChatGPT o1~\cite{achiam2023gpt}, following the procedure described in Sec.~\ref{method:scene_graph_generation}.
Using the scene graphs generated from each method, we performed place recognition and observed a significant difference in performance. The OSM-based method achieved a top-10 Retrieval Recall of 0.86, whereas the human-annotated and LLM-processed descriptions achieved 0.43. We report the top-10 metric for this comparison to provide a more meaningful numerical evaluation, since the number of correct matches within top-1, top-3, and top-5 was limited for the human-generated descriptions.
This performance gap is primarily attributed to differences in object descriptions. For example, a \textit{parking lot} in the OSM-based text may be referred to as a \textit{roadside parking lot} in the LLM-processed version. Additionally, certain objects visible in street view imagery, such as \textit{sidewalks}, were not present in the OSM data. Including these additional elements in the scene graphs was found to reduce the accuracy of place recognition.
\section{Conclusion}
In this letter, we proposed GOTPR which matches scene descriptions with OSM-based scene graphs to identify the current scene. GOTPR reduces the storage burden compared to traditional methods using point cloud maps by storing map data as scene graphs. Additionally, it leverages a vectorDB based on embedding vectors to identify matching candidates from the entire OSM scene graph database and performs scene retrieval within these candidates. This approach reduces retrieval time, enhancing the potential for real-world robotic applications. Lastly, the use of publicly available map data, such as OSM, which includes global outdoor information, provides advantages in terms of scalability and accessibility.\\
\indent This paper assumes that the text scene graph is a subgraph of the OSM scene graph, which simplifies the problem and enhances scene retrieval performance. However, this approach struggles with scenarios where the structures of the two graphs differ. To address this, future work will explore spatial embeddings to assess semantic similarity between the text and OSM scene graphs, moving beyond subgraph-based comparisons.

\bibliographystyle{IEEEtran}
\bibliography{main}
\end{document}